\documentclass[journal, transmag]{IEEEtran}
\ifCLASSINFOpdf
   \usepackage[pdftex]{graphicx}
\else
\fi
\usepackage{amsmath}
\usepackage{color}
\usepackage{comment}
\usepackage{multirow}
\usepackage{algorithm}
\usepackage{algpseudocode}
\graphicspath{{./pdf/}{./jpg/}{./png/}}
\ifCLASSOPTIONcompsoc
  \usepackage[caption=false,font=normalsize,labelfont=sf,textfont=sf]{subfig}
\else
  \usepackage[caption=false,font=footnotesize]{subfig}
\fi
\DeclareMathOperator*{\argmin}{arg\,min}
\newif\ifcolor
\colortrue

\begin{document}
\title{Prototype-based interpretation of the functionality \\ of neurons in winner-take-all neural networks}
\author{\IEEEauthorblockN{Ramin Zarei Sabzevar\IEEEauthorrefmark{1},
Kamaledin Ghiasi-Shirazi\IEEEauthorrefmark{1}, and
Ahad Harati\IEEEauthorrefmark{1}}
\IEEEauthorblockA{\IEEEauthorrefmark{1}Department of Computer Engineering, Ferdowsi University of Mashhad (FUM), Mashhad, IRAN}
\thanks{
Corresponding author: Kamaledin Ghiasi-Shirazi (email: k.ghiasi@um.ac.ir).}}
\markboth{Journal of \LaTeX\ Class Files,~Vol.~14, No.~8, August~2020}%
{Shell \MakeLowercase{\textit{et al.}}: Bare Demo of IEEEtran.cls for IEEE Transactions on Magnetics Journals}
\IEEEtitleabstractindextext{%
\begin{abstract}
Prototype-based learning (PbL) using a winner-take-all (WTA) network based on minimum Euclidean distance (ED-WTA) is an intuitive approach to multiclass classification.
By constructing meaningful class centers, PbL provides higher interpretability and generalization than hyperplane-based learning (HbL) methods based on maximum Inner Product (IP-WTA) and can efficiently detect and reject samples that do not belong to any classes. 
In this paper, we first prove the equivalence of IP-WTA and ED-WTA from a representational point of view.
Then, we show that naively using this equivalence leads to unintuitive ED-WTA networks in which the centers have high distances to data that they represent.
We propose $\pm$ED-WTA which models each neuron with two prototypes: one positive prototype representing samples that are modeled by this neuron and a negative prototype representing the samples that are erroneously won by that neuron during training.
We propose a novel training algorithm for the $\pm$ED-WTA network, which cleverly switches between updating the positive and negative prototypes and is essential to the emergence of interpretable prototypes.
Unexpectedly, we observed that the negative prototype of each neuron is indistinguishably similar to the positive one.
The rationale behind this observation is that the training data that are mistaken with a prototype are indeed similar to it.
The main finding of this paper is this interpretation of the functionality of neurons as computing the difference between the distances to a positive and a negative prototype, which is in agreement with the BCM theory.
In our experiments, we show that the proposed $\pm$ED-WTA method constructs highly interpretable prototypes that can be successfully used for detecting outlier and adversarial examples.
\end{abstract}
}
\maketitle
\IEEEdisplaynontitleabstractindextext
\IEEEpeerreviewmaketitle

\section{Introduction}
\IEEEPARstart{A}{} natural method for solving multiclass classification problems is prototype-based learning (PbL) in which one allocates several prototypes to each class and assigns an input data to the class of the nearest prototype, as measured by a distance function. 
This kind of learning has several advantages over hyperplane-based learning (HbL) methods such as Fisher linear discriminant analysis (LDA), support vector machines (SVM), and neural networks. 
The first advantage of PbL is the high interpretability: the input is assigned to a class since it is similar to one of its prototypes. 
Interpretability of learning machines is of paramount importance in some areas such as healthcare, criminal justice, and finance, and has recently become an important subject of research \cite{agarwal2020neural, maji2020exclusion, de2020deep, rudin2019stop, li2019application, nori2019interpretml}.
The second advantage of PbL is that it is suitable for few-shot learning \cite{fewshotNN,fewshotNips}.
Usually, the average of a few similar samples can be used as a prototype with high generalization.
The third advantage of PbL is that, since it is based on a distance function, it provides a high-quality reject option \cite{robustCNN}.
Although margin-based methods like \cite{svmReject, Platt99} provide a reject option based on the distance to margin, these methods can only reject ambiguous cases that necessarily belong to one of the classes.
Since there is no reason that a clutter input, which belongs to none of the classes, would fall near the margin of a hyperplane, HbL methods cannot reject them. 
On the other hand, since it is unlikely that a clutter input would fall in a small distance to a prototype, PbL methods can easily reject them.
Hence, measuring similarity based on a distance function, and not a general similarity measure like inner-product, is at the core of PbL. 
Consequently, PbL does not include hyperplane-based WTA classifiers such as multiclass SVM \cite{multiclassSVM} and neural networks with a softmax at their last layer.
Despite the aforementioned benefits of PbL, most commonly used classification systems, like SVM or neural networks, are HbL methods. 
We believe that this is because of two reasons: first, the recognition rates of the existing PbL methods are inferior to HbL methods like deep neural networks.
Second, the relationship between these two classes of algorithms is not studied in enough depth to offer necessary improvements from one area to the other.

Few efforts have been made to explore the relationship between HbL and PbL.
Graf et. al. \cite{prototypeBinary} demonstrated the relationship between PbL and Fisher LDA, SVM, and RVM classifiers, in binary classification.
Chang et. al. \cite{apl} examined the PbL problem from a learning theory perspective.
Oyedotun and Khashman \cite{piemnn} categorized learning algorithms into two disjoint groups of prototype learning and adaptive learning. They suggested using a combination of these two learning methods in a learning system to exploit the benefits of both.
In this paper, we combine the benefits of PbL and HbL by introducing a prototype-based interpretation of the functionality of neurons.
We consider WTA structures based on the minimum Euclidean Distance (ED) to a set of centers and the maximum Inner Product (IP) with a set of weight vectors and call them ED-WTA and IP-WTA, respectively. 
IP-WTA is an HbL method and ED-WTA is a PbL method.
We prove that ED-WTA and IP-WTA models have the same representational power and that each ED-WTA is equivalent to an IP-WTA, and vice versa.
We use this equivalence to find an ED-WTA network which is equivalent to a trained IP-WTA network. 
Besides, we propose an amended competitive cross-entropy algorithm\cite{cce} for training ED-WTA networks from scratch. 
However, we found out that the centers of the obtained ED-WTA network deviate from the input data and can not be considered as prototypes of data.
The reason is that, during discriminative training of WTA networks, weights/centers not only are pulled towards the training samples but also are repelled from them. 
Therefore, each weight/center is the difference between two components, one representing a combination of samples that are added to it and the other representing a combination of samples that are subtracted from it. 

We first propose a generalized HbL model, called $\pm$IP-WTA, in which the weight vector of a neuron is represented as the difference of two weight vectors: one representing samples that are added to it, and the other representing samples that are subtracted from it.
We then propose a generalized PbL model, called $\pm$ED-WTA, in which the weight vector of a neuron is modeled with two prototypes, a positive prototype representing the samples that belong to the neuron and a negative prototype representing the samples that are erroneously assigned to the neuron during training.
Interestingly, the computation of neurons of $\pm$ED-WTA with positive and negative centers simplify to IP. 
Modeling the weight vector of a neuron with two prototypes is of paramount importance as it gathers PbL and HbL together in artificial neural networks.
Given any discriminative training algorithm for IP-WTA networks, we propose a training algorithm for $\pm$IP-WTA and $\pm$ED-WTA that properly shares out the training updates between the positive and negative components such that the networks are kept equivalent to the original IP-WTA during training.  
The visualization of these positive and negative weights and centers revealed an interesting and somewhat unexpected fact. 
We observed that the negative prototype of each neuron, which had been created solely by input data from other classes, was indistinguishably similar to the positive prototype of that neuron. 
The rationale behind this observation is that the training data that are mistaken with a prototype are indeed similar to it.
This interpretation of the functionality of IP neurons as computing the difference between the similarities to a positive and a negative prototype, which themselves are extremely similar, is one of the main findings of this paper.
This finding is in agreement with the BCM theory \cite{bcm} which states that biological neurons discriminate between those input stimuli that excite the postsynaptic neuron strongly (here, those comprising the positive prototype) and those input stimuli that excite the postsynaptic neuron weakly (here, input patterns from other classes which are very similar to the positive prototype) \cite{hopfield}. 
Input stimuli that do not excite the postsynaptic neuron (here, those which are not similar to the positive prototype at all) do not affect the synaptic strength to that neuron.

The paper proceeds as follows. 
First, in section~\ref{sec:WTA}, we formally define IP-WTA and ED-WTA networks. 
Then, in section~\ref{sec:Equivalence_IP_ED}, we show that corresponding to each IP-WTA there exists an ED-WTA with identical functionality and vice versa. 
However, considering that the centers should be close to the input data that they represent, in subsection~\ref{sec:iterativeAlgorithm}, we propose an iterative algorithm that uses the weights of an IP-WTA network to obtain the centers of an equivalent ED-WTA whose centers are close to the input data.
In section~\ref{sec:ed-cce}, we propose a tailored version of the Competitive Cross-Entropy (CCE) loss \cite{cce} for training ED-WTA from scratch.
In section~\ref{sec:PN_prototype}, we introduce $\pm$IP-WTA and $\pm$ED-WTA networks in which two sets of positive and negative weights/centers are used in place of the original ones. 
In section~\ref{sec:Exp}, we report our experimental results on MNIST and ORL dataset.
Furthermore, we show the usefulness of the proposed $\pm$ED-WTA  model for rejecting outlier and adversarial inputs. We conclude the paper in section~\ref{sec:Conclusions}.
\section{Inner-product and Euclidean-distance Winner-Take-All networks}\label{sec:WTA}
Two commonly used criteria for measuring the similarity between an input data and a prototype are Euclidean distance and inner product.
Conceptually, the notion of similarity is best conveyed by Euclidean distance (ED) while the inner product (IP) operation has benefits in terms of faster computation and biological plausibility.
Most artificial neural networks use the additive neuron model which is inspired by the biological neurons in the brain. In this model, a single neuron computes a weighted sum of the activities of the presynaptic neurons plus a bias term that models the \textit{threshold potential} in the biological neurons. 
In this paper, we refer to this model as the IP model of the neuron, since the output is computed as the inner product of the input vector with the weight vector.
In this paper, we consider WTA networks with several neurons for each class, in which, the network output is determined by the label of the neuron whose weights/centers are most similar (based on either ED or IP) to the input data.
Consider a multiclass classification problem with $K$ classes and assume that the output layer of the network has $ M \geq K $ neurons. For each $k \in \lbrace 1,\dots,K\rbrace $, let $O_{k}$ be the set of indices of output neurons that are assigned to class $k$. If we use IP to measure the similarity of an input vector $x$ to the weight vectors $w_1,\dots,w_M$, then the label predicted by the network is calculated as: 
\begin{equation}
{\mathop{{argmax}}_{k\in \left\{1,\dots ,K\right\}} {\mathop{{max}}_{j\in O_k} w^T_jx}}.
\label{eq:IP}
\end{equation}
If we assume that the above-mentioned IP is applied to the augmented input $[x, 1]$ and the weights connecting the appended input $1$ to the output neurons are $b_1,\dots, b_M$, then we obtain IP-based winner-take-all (IP-WTA) which computes the output as:
\begin{equation}
y^{IP}={\mathop{{argmax}}_{k\in \left\{1,\dots ,K\right\}} {\mathop{{max}}_{j\in O_k} w^T_jx+b_j}}.
\label{eq:IP_WTA}
\end{equation}
Similarly, in the case of ED with centers $c_1, \dots,c_M$, the predicted label is computed as:
\begin{equation}
{\mathop{{argm}in}_{k\in \left\{1,\dots ,K\right\}} {\mathop{{min}}_{j\in O_k} {\left\|x-c_j\right\|}^2\ }}.
\label{eq:ED}
\end{equation}
If the input is augmented as $[x, 0]$ and the entries in the centers that correspond to the $0$ input are $d_1,\dots, d_M$, then we obtain an ED-based winner-take-all (ED-WTA)network which computes the output as:
\begin{equation}
y^{ED}={\mathop{{argm}in}_{k\in \left\{1,\dots ,K\right\}} {\mathop{{min}}_{j\in O_k} {\left\|x-c_j\right\|}^2+d^2_j\ }}.
\label{eq:ED_WTA}
\end{equation}
In practice, IP-WTA networks such as linear SVM are more commonly used than ED-WTA networks like LVQ. In deep neural networks for classification the output class is also usually computed based on IP-WTA using a softmax layer.
Despite the widespread use of IP-WTA in neural networks, ED-WTA has the advantage that its centers can be interpreted as prototypes of classes.
An interesting question is whether the prototypes of an ED-WTA can be obtained from the weights of a trained IP-WTA?
If it would be possible, then we could match each neuron with a prototype, obtaining both a prototype-based interpretation and a high recognition rate.
\section{Equivalence of IP-WTA and ED-WTA networks}
\label{sec:Equivalence_IP_ED}
In this section, we show that the modeling capabilities of IP-WTA and ED-WTA networks are the same. To this end, we show that there is an equivalent IP-WTA corresponding to each ED-WTA and vice versa.
\subsection{From ED-WTA to IP-WTA}
For WTA networks with a single neuron for each class, \cite{martin} showed that there is an equivalent IP-WTA corresponding to each ED-WTA. The result can be trivially extended to the case of multiple neurons for each class. We will state this result for the sake of completeness.
We show that for any ED-WTA network with centers $c_1,...,c_M$ and biases $d_1,...d_M$, there exists an IP-WTA network with weights $w_1,...,w_M$ and biases $b_1, ... , b_M$ such that, for any input data, the winning neurons of both models are the same.
Starting from \eqref{eq:ED_WTA} we have:
\begin{equation}
\begin{aligned}
y^{ED} &= \argmin \limits_{k\in\{1,\dots ,K\}} \min\limits_{j\in O_{k}} \parallel c_j - x \parallel^2 + d_j^2 \\
 &= \argmin \limits_{k\in\{1,\dots ,K\}} \min\limits_{j\in O_{k}} (c_j - x)^T(c_j - x) + d_j^2 \\
&= \argmin \limits_{k\in\{1,\dots ,K\}} \min\limits_{j\in O_{k}} (c_j^Tc_j-2c_j^Tx+x^Tx+d_j^2) .\label{eq:ED2IP}
\end{aligned}
\end{equation}
Considering that for any input data $x$ the value $x^T x$ is fixed, we can eliminate it and obtain:
\begin{equation}
\begin{aligned}
y^{ED} &= {\mathop{{argm}in}_{k\in \left\{1,\dots ,K\right\}} {\mathop{{min}}_{j\in O_k} \left({c_j}^Tc_j{-2}{c_j}^Tx+d^2_j\right)\ }\ \ }\\
&={\mathop{{argmax}}_{k\in \left\{1,\dots ,K\right\}} {\mathop{{max}}_{j\in O_k} \left({c_j}^Tx-\frac{{1}}{{2}}\left({c_j}^Tc_j+d^2_j\right)\right)}}.
\label{eq:ED_min_max_forms}
\end{aligned}
\end{equation}
Hence, if we choose the parameters of the IP-WTA model as $w_j=c_j$ and $b_j=-\frac{1}{2}{c_j}^T c_j-\frac{1}{2}d^2_j$, then the two models become equivalent.
\subsection{From IP-WTA to ED-WTA}
Suppose we have an IP-WTA network with weights $w_1,...w_M$ and biases $b_1,...,b_M$.
It is sufficient to show the existence of an ED-WTA network with some parameters $c_1,\dots ,c_M$ and $d_1,\dots ,d_M$ such  that the winning neurons of both networks are the same for all inputs $x$, i.e.
\begin{equation}
{\mathop{{argmax}}_{k\in \left\{1,\dots ,K\right\}} {\mathop{{max}}_{j\in O_k} w^T_jx+b_j}}={\mathop{{argm}in}_{k\in \left\{1,\dots ,K\right\}} {\mathop{{min}}_{j\in O_k} {\left\|c_j-x\right\|}^{{2}}+d^2_j\ }}.
\label{eq:IP2ED}
\end{equation}
Using \eqref{eq:ED_min_max_forms}, it is enough to prove that:
\begin{equation}
{\mathop{{argmax}}_{k\in \left\{1,\dots ,K\right\}} {\mathop{{max}}_{j\in O_k} w^T_jx+b_j}}
={\mathop{{argmax}}_{k\in \left\{1,\dots ,K\right\}} {\mathop{{max}}_{j\in O_k} {c_j}^Tx+e_j}},
\label{eq:IP_ED_EQUIV_GOAL}
\end{equation}
where $e_j=-\frac{{1}}{{2}}\left({c_j}^Tc_j+d^2_j\right)$, for $j=1,...,M$. 
For \eqref{eq:IP_ED_EQUIV_GOAL} to be an identity, it suffices to exist a positive coefficient $\alpha>0$ and another constant $\gamma$ such that for every $j\in \{\ 1,\dots ,M\} $ and any input $x$
\begin{equation}
\alpha \left(w^T_jx+b_j\right)-\gamma =c^T_jx+e_j .
\label{eq:9}
\end{equation}
For $j\in \{\ 1,\dots ,M\}$, the values $w_j$ and $b_j$ are given from IP-WTA and the values of $c_j$, $d_j$, $\alpha$, and $\gamma$ must be chosen in such a way that the above equation holds.
This choice is not unique and there are different solutions for each choice of $\alpha >0$. 
The above equation holds if and only if for every $j\in \{\ 1,\dots ,M\}\ $, we have ${\alpha w}_j=c_j$ and $\alpha b_j-\gamma =e_j$ that leads to
\begin{equation}
\alpha b_j-\gamma =-\frac{{1}}{{2}}\left({c_j}^Tc_j+d^2_j\right)=-\frac{1}{2}\left({\alpha }^2{w_j}^Tw_j+{d_j}^2\right).
\end{equation}
Therefore, for each choice of $\alpha >0$ and $\gamma$, $d_j$ is given by:
\begin{equation}
d_j=\pm \sqrt{-{\alpha }^2w^T_jw_j-2\alpha b_j+2\gamma }
\label{eq:10}
\end{equation}
In order to ensure that the expression under radical is not negative, it suffices to choose $\gamma$ greater than $\gamma_0=\frac{1}{2}\mathop{{max}}_{j}({\alpha }^2{w_j}^Tw_j+2\alpha b_j)$.
Therefore, any choice $\alpha>0$ and $\gamma\ge \gamma_0$ yields an ED-WTA with $c_j=\alpha w_j$ and $d_j$ as given by \eqref{eq:10} that is equivalent to an IP-WTA with weights $w_j$ and biases $b_j$, where $j=1,...,M$. 
\subsection{Finding a natural equivalent ED-WTA}\label{sec:iterativeAlgorithm}
Assume that the weights $w_1, ..., w_M$ are obtained by training an IP-WTA network.
We want to find the centers $c_1, ..., c_M$ in such a way that, firstly, the ED-WTA network has the same performance as the original IP-WTA network and, secondly, the centers $c_1, ..., c_M$ are close to the input data that they represent.
Let the centers $c_j$ and the weights $w_j$ be related by $c_j=\alpha w_j+u$, where $u$ is fixed and does not depend on $j$. 
For each input data $x$, let $q(x)$ be the index of the winning neuron in IP-WTA which is mathematically defined as:
\begin{equation}
q\left(x\right)={\mathop{{argmax}}_{1\le j\le M} w^T_jx+b_j\ }.
\end{equation}
Consider the following error function, measuring the sum of the squared distances between input data and the nearest center:
\begin{eqnarray}
\begin{aligned}
E(\alpha ,u)&=\sum_{x\in \mathcal{D}}{{\left\|x-c_{q\left(x\right)}\right\|}^2}=\sum_{x\in \mathcal{D}}{{\left\|x-(\alpha w_{q\left(x\right)}+u)\right\|}^2} \\
&=\sum_{x\in \mathcal{D}}\left[{\alpha }^2{\left\|w_{q\left(x\right)}\right\|}^2+{\left\|u\right\|}^2-2\alpha w^T_{q\left(x\right)}x\right.\\ 
&\left.-2u^Tx+2\alpha u^Tw_{q\left(x\right)}\right]+const,
\end{aligned}
\label{eq:closeness_loss}
\end{eqnarray}
where $\mathcal{D}$ is the set of training samples.
To minimize $E(\alpha ,u)$, we set the partial derivatives with respect to the parameters $\alpha$ and $u$ equal to zero:
\begin{gather*}
\begin{aligned}
\frac{\partial E}{\partial \alpha }&=\sum_{x\in T}{2\alpha {\left\|w_{q\left(x\right)}\right\|}^2-2{w_{q\left(x\right)}}^Tx+2u^Tw_{q\left(x\right)}}=0 \\ 
\frac{\partial E}{\partial u}&=\sum_{x\in T}{2u-2x+2\alpha w_{q\left(x\right)}}=0,
\end{aligned}
\end{gather*}
and obtain the following fixed-point equations:
\begin{eqnarray}
\begin{aligned}
\alpha &=\frac{\sum_{x\in \mathcal{D}}{({w_{q\left(x\right)}}^Tx-u^Tw_{q\left(x\right)})}}{\sum_{x\in \mathcal{D}}{{\left\|w_{q\left(x\right)}\right\|}^2}} \\ 
u&=\frac{1}{|\mathcal{D}|}\left(\sum_{x\in \mathcal{D}}{x-\alpha w_{q\left(x\right)}}\right).
\end{aligned}
\label{eq:alpha&u_values}
\end{eqnarray}
We start with $\alpha=0$ and alternate between updating $u$ and $\alpha$ using \eqref{eq:alpha&u_values}. Note that for all choices of $\alpha$ and $u$, the ED-WTA network with centers $c_j=\alpha w_j + u$ and ED biases $d_j = \pm\sqrt{c_j^T c_j -2\alpha b_j + 2\gamma}$ is equivalent to the original IP-WTA network.
Since our main goal is to obtain prototypes which are similar to input data, the natural choice for the ED biases $d_1, ..., d_M$ is zero, as they correspond to a feature with value zero in the augmented input vectors $[x,0]$.
Thus, at the end of the above-mentioned fixed-point optimization, we remove the parameters $d_1, ..., d_M$ from ED-WTA which breaks the equivalence between the IP-WTA and ED-WTA networks, and considerably deteriorates the accuracy of the ED-WTA network.
To take back the lost accuracy, in the next section, we introduce a variant of the CCE loss for optimizing ED-WTA networks.
\section{Training ED-WTA with CCE loss}\label{sec:ed-cce}
The CE (Cross Entropy) loss is ubiquitously used for training neural networks in multiclass classification problems. Competitive Cross-Entropy (CCE) is a generalization for the case of more than one neuron per class \cite{cce}. Here, we tailor CCE loss for training ED-WTA networks. In contrast to IP-WTA in which the winning neuron is the neuron with maximum activity, in ED-WTA networks the winning neuron is the least active one, the one with the lowest distance to the learnt prototype. 
To make the equations of IP-WTA and ED-WTA symmetric, we modify the functionality of ED neurons slightly with formula\footnote{For a generalization of IP and ED neurons, see \textit{L2 family of generalized convolution operators} in \cite{gcnn} which is a family of similarity measures that smoothly transform from IP to ED.}:
\begin{equation}
z_j{=-}\frac{1}{2}{\left\|x-c_j\right\|}^2\ \ \ \ for\ j=1,\dots ,M
\label{eq:ED_neurons}
\end{equation}
Now, consider an ED-WTA network with M output neurons and, for $k=1,...,K$, let $O_k$ denote the set of indices of output neurons that are assigned to class $k$. 
For $j=1,...,M$, let $z_j$ be the value of the output ED neuron $j$. 
To convert the output of the network to a probability distribution, we apply a softmax function with some parameter $\beta$ to $z$ and obtain the probability distribution $y$ with: 
\begin{equation}
y_j=\frac{e^{\beta z_j}}{\sum_{i=1}^{M}{e^{\beta z_i}}},\ \ \ \ for\ j=1,\dots ,M.
\label{eq:ED_Sotfmax}
\end{equation}
As suggested in \cite{cce}, the target distribution $\tau$ for each sample input is constructed by competition among the neurons which belong to the true class (say k):
\begin{equation}
{\tau }_j = \begin{cases}
\frac{e^{\beta z_j}}{\sum_{i\in O_k}{e^{\beta z_i}}} &j\in O_k\\
0 &otherwise.
\end{cases}
\label{eq:true_lable_distribution}
\end{equation}
The instantaneous CCE loss for a single training sample is defined as the cross entropy between the target distribution and the output distribution:
\begin{equation}
E^{CCE}=-\sum^M_{j=1}{{\tau}_{j}ln\left({y}_{j}\right)} .
\label{eq:CCE_loss}
\end{equation}
The inclusion of the parameter $\beta$ in the softmax operation for ED-WTA is crucial. 
The reason is that unless the input values $z$ to the softmax layer are in an appropriate range, the gradient backpropagated by the softmax layer would vanish.
In IP-WTA, the weights themselves can be adjusted to bring $z$ to an appropriate range. However in ED-WTA, the centers should be kept close to the input data and therefore an extra parameter $\beta$ is needed to put the values of $z$ in an appropriate range for the softmax operation. 
The parameter $\beta$ is trained, like all the parameters of the network, with gradient descent.
The gradient of the error function $E^{CCE}$ with respect to the centers $c_1,...,c_M$ and the parameter $\beta$ are:
\begin{eqnarray}
\begin{aligned}
\frac{\partial E^{CCE}}{\partial c_j}&=\frac{\partial E^{CCE}}{\partial z_j}\frac{\partial z_j}{\partial c_j}=\beta \left(y_j-{\tau }_j\right)\left(x-c_j\right)\\
\frac{\partial E^{CCE}}{\partial \beta }&=-\frac{1}{2}\sum^{M}_{j=1}{\left(y_j-{\tau }_j\right){\left\|x-c_j\right\|}^2} .
\label{eq:gradient_CCE_loss}
\end{aligned}
\end{eqnarray}
Using a learning rate of $\mu$, the formulas for updating the centers $c_1,...,c_M$ and the parameter $\beta$ by stochastic gradient descent are:
\begin{align}
c^{new}_j&=c^{old}_j-\mu \beta \left(y_j-{\tau }_j\right)\left(x-c_j\right) \label{eq:updating_centers}\\
{\beta }^{new}&={\beta }^{old}+\frac{1}{2}\mu \sum^{M }_{j=1}{\left(y_j-{\tau }_j\right){\left\|x-c_j\right\|}^2}
\label{eq:updating_beta}
\end{align}
Remember that we assumed that the true label is $k$.
For output neurons of the true class, i.e. when $j\in O_k$, we have:
\begin{eqnarray}
y_j=\frac{e^{\beta z_j}}{\sum_{i{\in }{{O}}_{{k}}}{e^{\beta z_i}}+\sum_{i{\notin }{{O}}_{{k}}}{e^{\beta z_i}}}\le \frac{e^{\beta z_j}}{\sum_{i{\in }{{O}}_{{k}}}{e^{\beta z_i}}}={\tau }_j.
\label{eq:y_tau_trueClass}
\end{eqnarray}
On the other hand, for neurons of other classes, i.e. when $j\notin O_k$, we have ${\tau }_j=0$ and therefore:
\begin{equation}
y_j=\frac{e^{\beta z_j}}{\sum^M_{i=1}{e^{\beta z_i}}}>{\tau }_j .
\label{eq:y_tau_otherClass}
\end{equation}
From \eqref{eq:y_tau_trueClass} and \eqref{eq:y_tau_otherClass}, it follows that in \eqref{eq:updating_centers}, the centers $c_j$ belonging to the true class move towards the input data while the centers $c_j$ which belong to other classes are repelled from it.
Stated another way, for each input sample, some positive multiples of it are added to the centers belonging to the true class and some positive multiples are subtracted from the centers of other classes.
Consequently, instead of being prototypes for input data, the centers of ED-WTA are differences of two prototypes of input data: one prototyping the data that belong to that center and another prototyping the data that are mistakenly won by that center.
\section{Explicit modeling of positive and negative prototypes}
\label{sec:PN_prototype}
As we saw in the previous section, the centers learned by CCE for 
ED-WTA were not prototypes for the data they represented.
The reason is that, during the discriminative training process,
while data belonging to a center appear with a positive sign in
the update relation, however, data belonging to other classes
will also contribute to the value of the center with a negative
sign. While we desire that the centers become approximately
equal to the mean of the data that they represent, nonetheless, the
discriminative training of the parameters causes this not to be
the case and the centers are a combination of the data with
positive and negative coefficients. 
Based on this observation, first of all, we split the parameters of each neuron into a positive parameter and a negative one. Then, we propose a new learning rule and apply all updates in which the input data appear with a positive sign to the positive parameter and those with a negative sign to the negative parameter. 
We apply this method to IP-WTA and ED-WTA models and call them $\pm$IP-WTA and $\pm$ED-WTA models, respectively. 
\subsection{$\pm$IP-WTA Model}
\label{sec:PN_IP_WTA}
Suppose that for each $j\in \{1,\dots ,M\}$, we split the weight vector $w_j$ in IP-WTA to $w_j=w^+_j-w^-_j$, where 
$w^+_j$ and $w^-_j$ , respectively, represent the contributions of data with positive and negative coefficients to the weight vector $w_j$.
We call this new model $\pm$IP-WTA and call $w^+$ and $w^-$ the positive and the negative weights, respectively \footnote{Please note that in general, when data takes negative values, $w^+_j$ and $w^-_j$ may also take negative values. So, don't confuse the contribution of data with a positive coefficient with the positivity of $w^+_j$.}.
For training the parameters, in the new formulation, always a positive coefficient of each input sample is accumulated with either the positive or the negative weights.
So, whenever the update rule of IP-WTA adds a positive coefficient of a training sample to $w_j$, we apply it to $w_j^+$, and whenever the update rule of IP-WTA adds a negative coefficient of a training sample to $w_j$, we apply this change with a positive coefficient to $w_j^-$.
Clearly, at each step, only one of the parameters $w_j^+$ or $w_j^-$ would be updated.
The updating rule of weights in CCE algorithm are:
\begin{eqnarray}
\begin{aligned}
{w_j}^{new}&={w_j}^{old}-\mu \left(y_j-{\tau }_j\right)x,\ \ \forall j\in O_k\\
{w_j}^{new}&={w_j}^{old}-\mu \left(y_j-0\right)x,\ \ \forall j\notin O_k .
\label{eq:updating_IP_weights}
\end{aligned}
\end{eqnarray}
We have already shown that in \eqref{eq:y_tau_trueClass} and \eqref{eq:y_tau_otherClass}, that the coefficient of $x$ is positive for $j\in O_k$, and negative for $j\notin O_k$. Therefore, updating of weights in $\pm$IP-WTA is as follows:
\begin{eqnarray}
\begin{aligned}
{{w_j}^+}^{new}&={{w_j}^+}^{old}+\mu \left({\tau }_j-y_j\right)x,\ \ \ \ \ \forall j\in O_k\\
{{w_j}^-}^{new}&={{w_j}^-}^{old}+\mu y_j x,\ \ {\ }\forall j\notin O_k ,
\label{eq:updating_IP_weights_in_new_struct}
\end{aligned}
\end{eqnarray}
The above formulation guarantees that both the positive and the negative weights will be combinations of data with positive coefficients.  
Although in $\pm$IP-WTA we obtain weights which are visually similar to the input data, the positive/negative weights are not prototypes of data since their magnitude is arbitrary.
To solve this drawback, in the next section, we apply this method to the ED-WTA model and obtain positive and negative prototypes.
\subsection{$\pm$ED-WTA Model}
\begin{figure}[!t]
\centering
\includegraphics[width=\columnwidth,keepaspectratio]{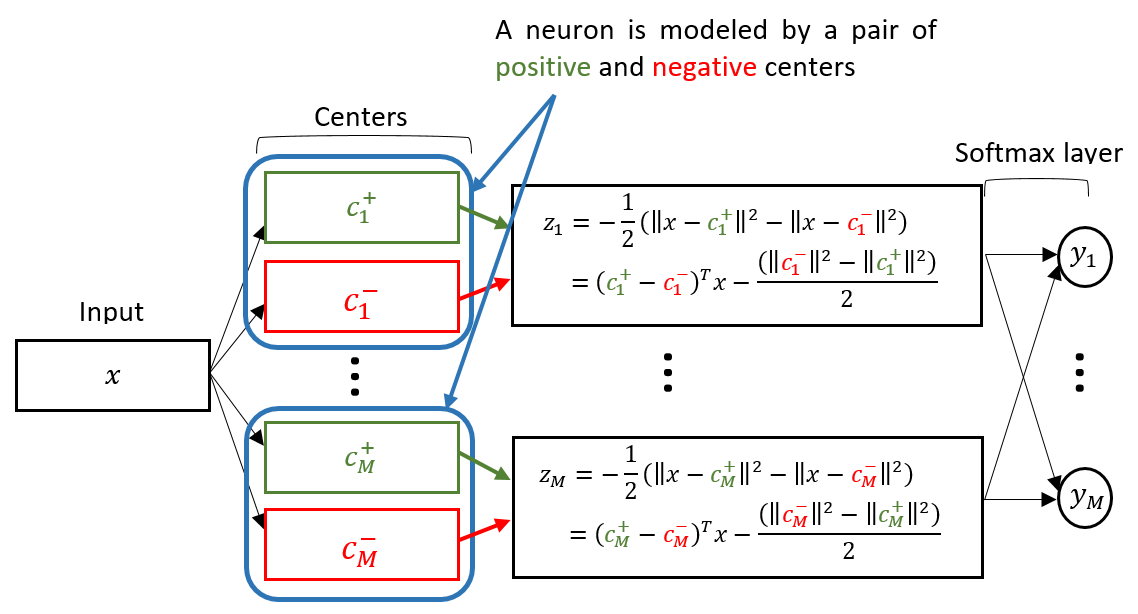}
\caption{The structure of a $\pm$ED-WTA network with positive and negative prototypes for each class. Each neuron is modeled by a positive and a negative prototype center. During training, we ensure that these centers are kept close, in Euclidean norm, to the input data that the they represent. In final usage, the whole functionality of the positive and negative centers is simplified to an ordinary IP-WTA network.}
\label{fig:new_struct}
\end{figure}

Naive splitting of the centers of ED-WTA results in the following new model:
\begin{equation}
z_j=-\frac{1 }{2}{\left\|x-\left(c^+_j-c^-_j\right)\right\|}^2,
\label{eq:naively_new_ED_model}
\end{equation}
where both $c_j^-$ and $c_j^+$ are obtained by accumulating training samples with non-negative coefficients. 
The problem with this model is that it favors a low distance between the input data and the difference of the positive center $c_j^+$ and the negative center $c_j^-$, not a low distance to the centers themselves. 
Therefore, we propose the following model for neurons and call it $\pm$ED-WTA:
\begin{equation}
z_j=-\frac{1 }{2}\left({\left\|x-c^+_j\right\|}^2-{\left\|x-c^-_j\right\|}^2\right).
\label{eq:new_ED_model}
\end{equation}
Considering that the new model is based on the distances between the centers $c_j^+$ and $c_j^-$ with the input data $x$, we expect that, by a proper training method, $c_j^+$ and $c_j^-$ would become prototypes of data. Interestingly, according to \eqref{eq:new_ED_model}, the $\pm$ED-WTA model simplifies to the IP-WTA model, as shown below:
\begin{equation}
z_j={\left(c^+_j-c^-_j\right)}^Tx-\frac{\left({\left\|c^-_j\right\|}^2-{\left\|c^+_j\right\|}^2\right)}{2}.
\label{eq:new_ED_model_as_IP}
\end{equation}
Please note that while the computation of $\pm$ED-WTA simplifies to IP-WTA, from the learning viewpoint, $\pm$ED-WTA differs considerably from IP-WTA.
Since the bias term in \eqref{eq:new_ED_model_as_IP} is written in terms of positive and negative centers, the gradient of the centers of $\pm$ED-WTA differ from the gradient of the weights in IP-WTA. 
Besides, since $c_j^+$ and $c_j^-$ should be prototypes of the data, the magnitude of $c_j^+ - c_j^-$ cannot be freely chosen and there is nothing to prevent the values $z_1,...,z_M$  in \eqref{eq:new_ED_model_as_IP} from falling into the saturated region of the softmax function in which the back-propagated error becomes approximately zero.
Therefore, for $\pm$ED-WTA there should be another parameter, say $\beta$, which adjusts the range of the input of the softmax function. 
The structure of a $\pm$ED-WTA network with explicit positive and negative centers is illustrated in Fig. \ref{fig:new_struct}. Note that according to \eqref{eq:new_ED_model} and \eqref{eq:new_ED_model_as_IP}, after training, $\pm$ED-WTA can be substituted with an ordinary IP-WTA network.  
Training the parameters of $\pm$ED-WTA can be based on any optimization algorithm like stochastic gradient descent, NAG, or Adam.
However, in order to achieve proper positive and negative prototypes, we ought to update the parameters in a more clever way. 
If we apply an optimization algorithm to the new formulation \eqref{eq:new_ED_model_as_IP} based on the positive centers $c_j^+$ and the negative centers $c_j^-$, then both the positive and the negative centers of all neurons are changed for each input data, violating the philosophy behind them.
Therefore, at each step, the positive centers $c_j^+$ belonging to the class of the input sample and the negative centers $c_j^-$ of other classes are updated.
Considering the CCE algorithm and assuming that the label of the current training data is $k$, we have: 
\begin{equation}
{\tau }_j = \begin{cases}
\frac{e^{\beta z_j}}{\sum_{i\in O_k}{e^{\beta z_i}}} &j\in O_k\\
0 &j\notin O_k.
\end{cases}
\end{equation}
The rules for updating the centers of $\pm$ED-WTA are:
\begin{eqnarray}
\begin{aligned}
{c_j^+}^{new}&={c_j^+}^{old}+\mu \beta \left({\tau }_j-y_j\right)\left(x-{c_j^+}^{old}\right),\ \ \ \forall j\in O_k\\
{{{c}_j}^-}^{new}&={{{c}_j}^-}^{old}+\mu \beta \left(y_j\right)\left({x-c_j^-}^{old}\right),\ \ \forall j\notin O_k
\label{eq:updating_ED_centers_new_model}
\end{aligned}
\end{eqnarray}
where $y$ is the probability distribution obtained by the softmax function according to \eqref{eq:ED_Sotfmax}. These rules always update the positive/negative centers towards the input data, making these centers prototypes of the input data.
In contrast to $\pm$IP-WTA in which the weights are updated with a multiple of input data, in $\pm$ED-WTA the centers are updated based on their distance to the input data.
Consequently, in contrast to $\pm$IP-WTA, in $\pm$ED-WTA the centers will remain close to the data that they represent. 
Therefore, we obtain positive and negative prototypes that are close to the input data in Euclidean norm. Our training algorithm for $\pm$ED-WTA is shown in Algorithm~\ref{euclid}. 
\begin{algorithm}
\caption{$\pm$ED-WTA training algorithm}\label{euclid}
	
\begin{algorithmic}[1]
	\Require Dataset $\mathcal{D}=\{(x_1,l_1),(x_2,l_2),...,(x_{|\mathcal{D}|},l_{|\mathcal{D}|})\}$
	\Require Learning rate $\mu$
	\State Initialize centers $c_j^+$ and $c_j^-$ for all $j=1,\dots ,M$
	\State Initialize parameter $\beta$
	\For {$iter=1:number \ of\ epochs$}
	\For {$i=1:|\mathcal{D}|$}
	\State Let $k=l_i$ denote the label of the current sample
	\For {$j=1,...,M$}
	\State {$z_j={\left(c^+_j-c^-_j\right)}^Tx_i-\frac{\left({\left\|c^-_j\right\|}^2-{\left\|c^+_j\right\|}^2\right)}{2}$}
	\State {$y_j = \frac{exp{(\beta z_j)}}{\sum^{M}_{i=1}{exp{(\beta z_i)}}}$}
	\EndFor	
	\State {$ \tau_j=\frac{exp{(\beta z_j)}}{\sum_{i}{exp{(\beta z_i)}}},\ \ \ \ \forall\ j\in O_{k}$}	
	\State {$ {c_j^+}\leftarrow {c_j^+}+\mu \beta \left({\tau }_j-y_j\right)\left(x_i-{c_j^+}\right),\ \ \ \forall j\in O_{k}$}
	\State {${{{c}_j}^-}\leftarrow{{{c}_j}^-}+\mu \beta \left(y_j\right)\left({x_i-c_j^-}\right),\ \ \forall j\notin O_{k}$}
	\State ${\beta }\leftarrow {\beta }-\mu \sum^{M}_{j=1}{\left(y_j-{\tau}_j\right)z_j}$
	\EndFor
	\EndFor
\end{algorithmic}
\end{algorithm}
\section{Experiments}
\label{sec:Exp}
In this section, we experimentally evaluate the proposed method on a set of tasks. 
In section \ref{sec:ExpOnMNIST}, we compare the recognition rate and the interpretability of the learned parameters of LVQ, IP-WTA, ED-WTA, $\pm$IP-WTA, and $\pm$ED-WTA on MNIST digit recognition dataset. 
Considering that our method of choice is $\pm$ED-WTA, and that the forward computation of $\pm$ED-WTA is equivalent to IP-WTA, we then focus on comparing $\pm$ED-WTA and IP-WTA.
Comparison with IP-WTA is also important since IP-WTA is the most serious competitor as it is widely used in the last layer of deep neural networks for multiclass classification.
To show the usefulness of the proposed $\pm$ED-WTA model, in section \ref{sec:reject}, we compare $\pm$ED-WTA and IP-WTA on the task of detecting outlier samples and, in section \ref{sec:adversarial}, we compare their robustness against adversarial examples.
\subsection{Comparing WTA models on MNIST}
\label{sec:ExpOnMNIST}
In this section, we compare various WTA models on the MNIST dataset, which is a standard dataset to evaluate neural networks on multiclass classification problems. 
This dataset contains $60,000$ training images and $10,000$ testing images from English handwriting digits with size $28\times28$.
We introduce the experimental setup in Section~\ref{sec:Experimental settings} and report the experimental results in Section~\ref{sec:Visualization}.
\subsubsection{Experimental settings}
\label{sec:Experimental settings}
In all experiments, the we set the number of output neurons for each class to 6 and the number of training epochs to 200.
We select the initial learning rate by  5-fold cross-validation from the set of values $\{10^{- 2}, ..., 10^2\}$. During training, we multiply the learning rate by $0.5$ after the completion of each epoch.
We initialize the centers of ED-WTA and positive prototypes of $\pm$ED-WTA by the clusters obtained using the K-means clustering algorithm.
We initialize the negative prototypes of $\pm$ED-WTA by adding a small random noise to the corresponding positive prototypes.
For training LVQ, the number of epochs and the neurons is the same as other algorithms, and other parameters are automatically selected by the LVQ-PAK package.
\subsubsection{Visualization of Weights}
\label{sec:Visualization}
\begin{figure*}[!t]
\centering
\ifcolor
\subfloat[]
{\includegraphics[width=0.2\textwidth,keepaspectratio]{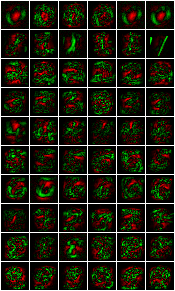}%
\label{fig_first_case}}
\hfil
\subfloat[]{\includegraphics[width=0.2\textwidth,keepaspectratio]{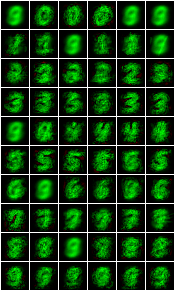}%
\label{fig_second_case}}
\hfil
\subfloat[]{\includegraphics[width=0.2\textwidth,keepaspectratio]{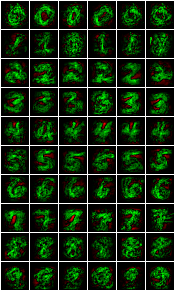}%
\label{fig_third_case}}
\hfil
\subfloat[]{\includegraphics[width=0.2\textwidth,keepaspectratio]{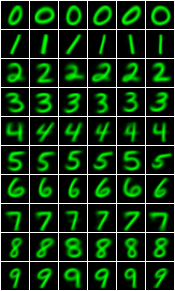}%
\label{fig_forth_case}}

\fi
\caption{Visualization of (a) trained IP-WTA weights, (b) the centers obtained by converting the weights of IP-WTA to ED-WTA using the algorithm of section \ref{sec:iterativeAlgorithm}, (c) the centers after dropping the biases of ED-WTA and continuing training with CCE for 200 epochs, (d) the centers obtained by the K-means algorithm. Positive and negative values are shown in green and red, respectively.}
\label{fig:ED_from_IP}
\ifcolor
\subfloat[]{\includegraphics[width=0.2\textwidth,keepaspectratio]{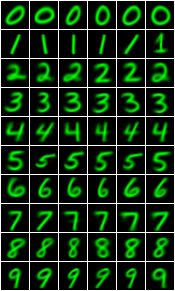}%
\label{fig_LVQ1.}}
\hfil
\subfloat[]{\includegraphics[width=0.2\textwidth,keepaspectratio]{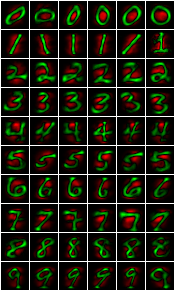}%
\label{fig_LVQ2.1}}
\hfil
\subfloat[]{\includegraphics[width=0.2\textwidth,keepaspectratio]{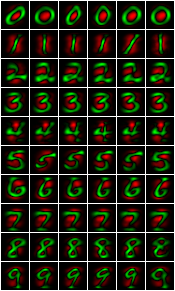}%
\label{fig_LVQ3}}
\hfil
\subfloat[]{\includegraphics[width=0.2\textwidth,keepaspectratio]{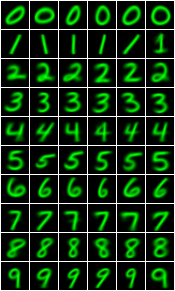}%
\label{fig_LVQ_functionality}}
\fi
\caption{Visualization of the weights of the (a) LVQ1, (b) LVQ2.1 and (c) LVQ3 algorithms. (d) The functionality of each output neuron after training LVQ2.1. Each cell shows the average of training images that maximally activate that neuron. Positive and negative values are shown in green and red, respectively.}
\ifcolor
\label{fig:LVQ}
\subfloat[]{\includegraphics[width=0.2\textwidth,keepaspectratio]{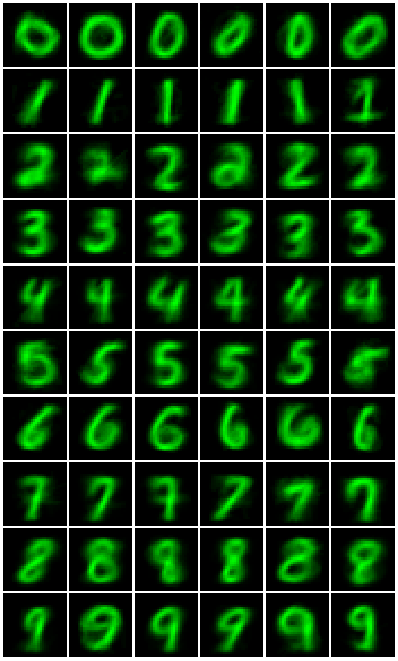}%
\label{fig_1_case}}
\hfil
\subfloat[]{\includegraphics[width=0.2\textwidth,keepaspectratio]{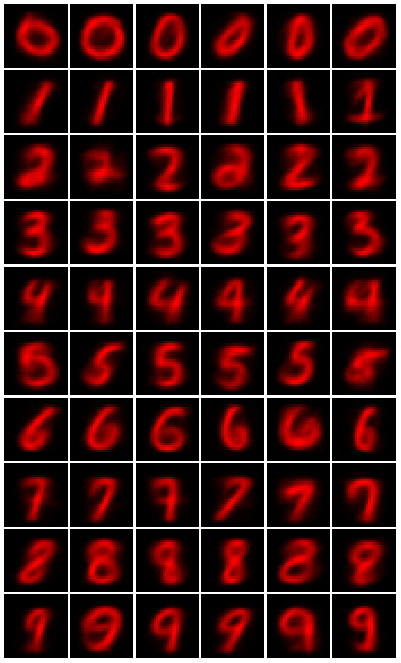}%
\label{fig_2_case}}
\hfil
\subfloat[]{\includegraphics[width=0.2\textwidth,keepaspectratio]{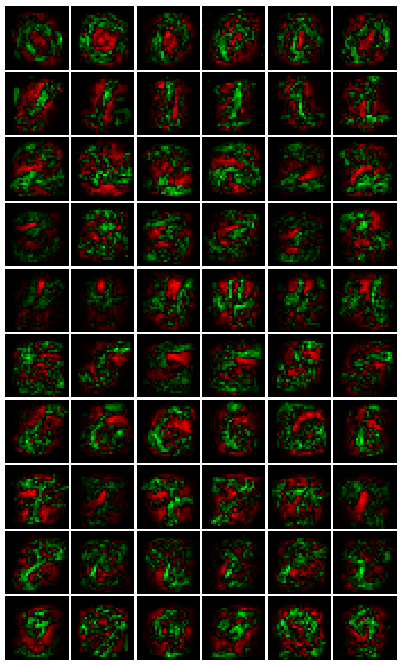}%
\label{fig_3_case}}
\fi
\caption{Visualization of the $\pm$ED-WTA. (a) Positive prototypes, (b) Negative prototypes, (c) Equivalent IP-WTA weights obtained by difference of positive and negative prototypes in (a) and (b). Positive and negative values are shown in green and red, respectively.}
\label{fig:Visual_PN_struct}
\end{figure*}

By visualizing the weights, one can obtain a clear view of what a WTA network has learned. 
\ifcolor
In all color images, positive values are shown in green and negative values are shown in red.
\fi
Fig. \ref{fig:ED_from_IP}.a shows the weights of an IP-WTA network trained on MNIST.
As can be seen, the weights did not emerge as prototypes of the data. The values of the weights of an IP-WTA network can be best described as a vote given by each pixel in favor/against the activity of each neuron.
The centers obtained by the algorithm of section \ref{sec:iterativeAlgorithm} applied to the MNIST dataset are shown in Fig \ref{fig:ED_from_IP}.b.
As can be seen, in comparison to the weights of IP-WTA, the centers of ED-WTA have become slightly more similar to the data of their associated classes. 
However, the obtained centers are still far from the prototypes of the data as can be understood by looking at the centers obtained by the k-means algorithm, shown in Fig.~\ref{fig:ED_from_IP}.d.
Besides, the accuracy decreased from 96.51\% in IP-WTA to 89.12\% when the bias terms of ED neurons had been dropped. 
By dropping the biases in ED-WTA and continuing training with CCE, the accuracy went up to 96.70\%, but the resulting prototypes became much worse as is shown in Fig. \ref{fig:ED_from_IP}.c.
Fig. \ref{fig:LVQ} shows the weights obtained by different versions of the LVQ algorithm. 
Although the centers in LVQ1 can be regarded as prototypes of data, LVQ1 has a considerably lower recognition rate of 90.78\%. 
On the other hand, the centers of LVQ2.1 and LVQ3 are not prototypical of data, while these algorithms obtain higher recognition rates of 94.04\% and 91.04\% respectively. 
Fig. \ref{fig:LVQ}.d shows the average of samples belonging to each neuron after training with LVQ2.1. It can be seen that the prototypes of data, as shown in Fig.~\ref{fig:LVQ}.d, significantly differ from the centers obtained by LVQ2.1, as shown in Fig.~\ref{fig:LVQ}.b.
The centers of Fig.~\ref{fig:LVQ}.b are created by a combination of training samples with positive and negative coefficients and even some pixels of the trained centers took negative values. Thus, the centers obtained by LVQ2.1  are not a proper representation for the corresponding classes.

Now, we report the results of the main proposal of this paper: $\pm$ED-WTA. 
Fig. \ref{fig:Visual_PN_struct}.a and Fig. \ref{fig:Visual_PN_struct}.b show the positive and the negative prototypes obtained by a $\pm$ED-WTA network trained on MNIST.
As desired, the positive prototypes for each class are prototypical for the corresponding data. Nevertheless, the astonishing result is that the negative centers have also developed to be representative of data of that class. 
This is particularly strange as these prototypes are merely obtained by samples from other classes.
However, this result becomes understandable by noticing that a negative prototype is formed by samples that are so similar to the positive prototype that are mistaken with it and wrongly classified.
Thus, it can be said that the negative prototypes are perturbed versions of their corresponding positive prototypes. 
Fig \ref{fig:Visual_PN_struct}.c shows the difference between the positive and negative prototypes, a visualization that is very similar to the weights of IP-WTA. Furthermore, $\pm$ED-WTA obtained a recognition rate of 96.95\%.
Table \ref{speci} summarizes the recognition rates of different methods studied in this paper on MNIST. As desired, $\pm$ED-WTA combines the high-accuracy of IP-based networks with the interpretability of ED-based models.
\begin{table}[!h]
\centering
\caption{The accuracy of different WTA models on MNIST.}
\resizebox{0.48\textwidth}{!}{
\begin{tabular}{c|c|c}
		Models    & Initialization method & Accuracy \\ \hline
        LVQ1      & LVQ-PAK initialization program & 90.78\% \\ \hline
        LVQ2.1    & LVQ-PAK initialization program & 94.04\% \\ \hline
        LVQ3      & LVQ-PAK initialization program & 91.04\% \\ \hline
		IP-WTA	  & Random & 96.51\% \\ \hline
\multirow{3}{*}{ED-WTA} & Random & 96.19\% \\
						& K-means & 96.23\% \\
                        & Method of section \ref{sec:iterativeAlgorithm} & 96.70\% \\ \hline
$\pm$IP-WTA & Random & 96.63\% \\ \hline
\multirow{2}{*}{$\pm$ED-WTA} & Random &  96.81\%\\
                             & K-means & \bf{96.95}\%
\end{tabular}%
\label{speci}
}
\end{table}
 
\subsection{Robustness against outlier inputs}
\label{sec:reject}
Although neural networks have achieved great success in pattern classification, they usually suffer from the vulnerability against out of class samples. So, when they are fed a sample from an unseen class, they still associate it to a known class, possibly with high confidence. 
In this section, we show the superiority of the proposed $\pm$ED-WTA  model against IP-WTA in detecting outlier data, which are data that belong to none of the classes.
For this propose, we train $\pm$ED-WTA and IP-WTA on the MNIST digits and evaluate their robustness by giving face samples from the ORL dataset \cite{samaria1994parameterisation_orl_dataset} during the testing phase. 
ORL face dataset contains 400 images captured from 40 distinct subjects. The images are grayscale of size $92\times112$ which we resized to $28\times 28$ for compatibility with the size of MNIST images.
We consider two distinct test sets: 
1) The test set of MNIST which consists of 10000 digits and 2) all 400 faces from the ORL dataset.
We evaluate the classifiers by two criteria: the true acceptance rate and the true rejection rate. 
The true acceptance rate is the ratio of the MNIST test samples that got accepted and the true rejection rate is the ratio of ORL samples that got rejected. 
Suppose that $k \in \{1,...,K\}$ is the predicted label (i.e. winning class) and that ${m} \in O_{k}$ is the index of the winning neuron in a $\pm$ED-WTA network. 
Considering that $\pm$ED-WTA simplifies to IP-WTA according to \eqref{eq:new_ED_model_as_IP}, the probability of the winning class $k$ considering IP neuron model is:
\begin{equation}
P^{IP}(x)={\frac{\sum_{j\in O_{k}}{{e^{z_j}}}}{\sum_{i=1}^{M}{e^{ z_i}}}}
\label{eq:CARS-IP}
\end{equation}

\begin{figure}[!t]
\centering
\includegraphics[width=0.49\textwidth,keepaspectratio]{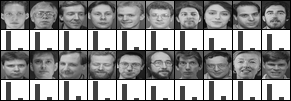}
\caption{Confidence measures of some outlier samples from the ORL dataset, computed by a $\pm$ED-WTA model trained on the MNIST dataset. The confidence measures $P^{IP}$ and  $P^{+ED}$ are drawn below each sample from left to right, respectively. 
}
\label{fig:rejection}
\end{figure}

\begin{figure}[!t]
\centering
\includegraphics[width=0.6\columnwidth,keepaspectratio]{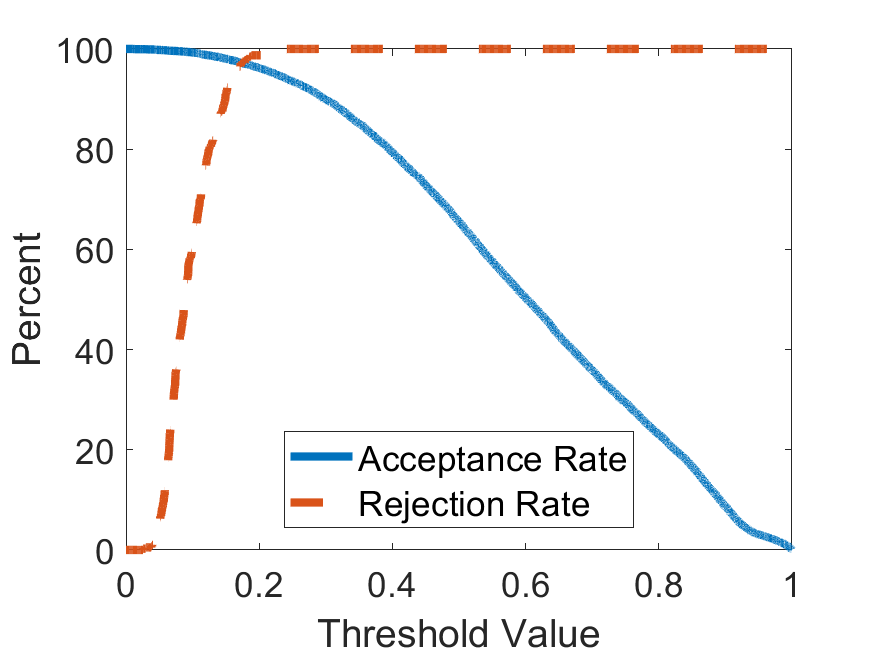}
\caption{Acceptance rate on the MNIST test set and the rejection rate on the ORL dataset for different threshold values of $P^{+ED}$.
}
\label{fig:rejectionDiag}
\end{figure}

To distinguish between outlier samples and input data that indeed belong to one of the classes, we consider the positive center of the winning neuron in $\pm$ED-WTA model and propose an additional confidence measure
\begin{equation}
P^{+ED}(x)={\frac{\sum_{j\in O_{k}}{{e^{-\frac{\beta}{2}{\left\|x-c^+_j\right\|}^2}}}}{\sum_{i=1}^{M}{e^{-\frac{\beta}{2}{\left\|x-c^+_i\right\|}^2}}}}
\label{eq:CARS}
\end{equation}
which is the probability that, based on distance to positive centers, sample $x$ belongs to the winning class $k$.
Fig. \ref{fig:rejection} shows several samples of the ORL dataset along with their confidence measures. 
While these samples have a high probability of $P^{IP}$ (in most cases, above 99\%), the $\pm$ED-WTA model assigns low probability of $P^{+ED}$ to them and easily detects them as outliers. 
Fig. \ref{fig:rejectionDiag} shows the acceptance rate on the MNIST test set and the rejection rate on the ORL dataset of the $\pm$ED-WTA model for different threshold values on $P^{+ED}$. For instance, by choosing the threshold value 0.19 for $P^{+ED}$, the acceptance and rejection rates will be 96.70\% and 98.75\% respectively. 
However, the equivalent IP-WTA, in which only the probability $P^{IP}$ can be utilized for rejection, achieves an acceptance rate of 77.95\% and a rejection rate of 64.50\% when we empirically chose the threshold value for $P^{IP}$ to maximize the product of the acceptance and rejection rates.
\subsection{Robustness against adversarial data}
\label{sec:adversarial}
\begin{figure}[!t]
\centering
\includegraphics[width=0.6\columnwidth,keepaspectratio]{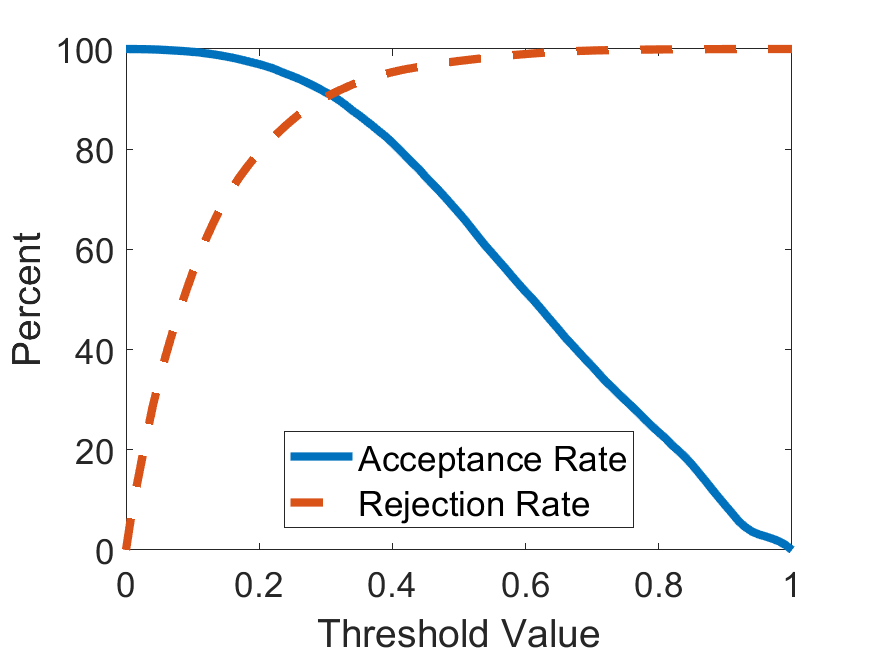}
\caption{Acceptance rate on the MNIST test set and the rejection rate on  100K adversarial data for different threshold values of $P^{+ED}$.
}
\label{fig:rejectionAdverDiag}
\end{figure}

Recent studies have shown that neural networks are vulnerable to adversarial examples \cite{adversarial}. Adversarial examples are generated by applying some small perturbations to real samples and can easily fool a neural network to predict wrong classes with high confidence.
In this section, we show the robustness of $\pm$ED-WTA against adversarial examples. For this purpose, we generate two types of adversarial data: 
\begin{itemize}
\item{Type-1:} Data that do not belong to any classes, while the model recognizes and assigns them to a class with a high probability.
\item{Type-2:} Data that are related to a certain class, but are classified incorrectly by the model with a high probability. 
\end{itemize}
To this end,  similar to the previous experiment, we first train a $\pm$ED-WTA model on the MNIST training data. By starting from 1000 pure noise images, we generate the Type-1 adversarial data in which each sample corresponds to one of the ten classes. For each sample, the values of pixels are updated by backpropagating errors from the equivalent IP-WTA network of the trained $\pm$ED-WTA model. 
To generate Type-2 adversarial data, we use 10000 MNIST test images and update their pixel values using backpropagation, while setting the target to one of the other nine classes. So, 10000 Type-1 data and 90000 Type-2 data are generated. Overall, we have 110000 testing data composed of 100000 adversarial data and the original 10000 MNIST test data. 
\begin{figure}[!t]
\centering
\subfloat[]{\includegraphics[width=0.49\textwidth,keepaspectratio]{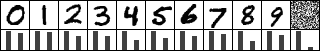}%
\label{Adverserial_case}}
\\
\subfloat[]{\includegraphics[width=0.49\textwidth,keepaspectratio]{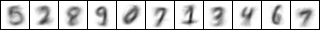}%
\label{Weight_case}}\\
\subfloat[]{\includegraphics[width=0.49\textwidth,keepaspectratio]{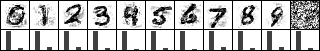}%
\label{Clean_case}}
\caption{The process of generating adversarial examples. (a) Some samples from the MNIST test set along with a pure-noise image.
(b) The positive centers of the neuron in $\pm$ED-WTA chosen as the target for generating an adversarial example. (c) The resulting adverserial examples. In (a) and (c), the probabilities $P^{IP}$ and $P^{+ED}$ are drawn from left to right below each sample. While $P^{IP}$ is high for both of the original and adversarial examples, $P^{+ED}$ is only high for the original digit images. The values of $P^{IP}$ and $P^{+ED}$ for the pure-noise image in (a) are very interesting. Since the pure-noise image is far from all hyperplanes, $P^{IP}$ confidentially accepts it as a digit. On the other hand, since the pure-noise image is far from all positive centers of $\pm$-ED-WTA, $P^{+ED}$ confidentially rejects it.
}
\label{fig:Adver}
\end{figure}
Fig. \ref{fig:Adver} shows the process of generating one Type-1 and ten Type-2 adversarial examples using the IP-WTA network obtained after training a $\pm$ED-WTA network on the MNIST dataset.
While the generated adversarial examples have high $P^{IP}$ probabilities, these samples can be easily rejected based on their low $P^{+ED}$ probabilities. 
Similar to the previous section, by choosing a threshold on output confidences, we obtained a detector for accepting the MNIST testing data and rejecting the adversarial examples.
Fig. \ref{fig:rejectionAdverDiag} shows the acceptance rate on the MNIST test set and the rejection rate on the adversarial data for the $\pm$ED-WTA model when choosing different threshold values for $P^{+ED}$. The $\pm$ED-WTA model achieved a 91.34\% acceptance rate and a 90.39\% rejection rate by choosing the threshold value 0.3 for $P^{+ED}$. However, we could not reject even 1\% of adversarial samples by using the probability $P^{IP}$, irrespective of the threshold value.
\section{Conclusions}
\label{sec:Conclusions}
In this paper, we investigated the problem of discovering a prototype-based interpretation of the functionality of neurons in winner-takes-all structures. 
After attempting different approaches, we concluded that, since neural networks are trained in a discriminative manner, the weights must be modeled as the difference of two prototypes, not one.
We proposed the $\pm$ED-WTA network in which each neuron is modeled by two prototypes based on the Euclidean distance: a positive prototype for modeling the data that are truly won by this neuron, and a negative prototype for those that are erroneously won. 
While training of $\pm$ED-WTA networks differs from ordinary neural networks, the resulting model of a trained $\pm$ED-WTA can be expressed as an ordinary WTA network based on IP.
This interpretation of the functionality of neurons as differentiating between those stimuli that are strongly similar to a prototype and those that are weakly similar to it is in agreement with the BCM theory.
Besides, as an application, we showed that the prototypes learned by $\pm$ED-WTA can be used for detecting outlier and adversarial examples.
\ifCLASSOPTIONcaptionsoff
  \newpage
\fi
\bibliographystyle{IEEEtran}

\begin{IEEEbiography}
[{\includegraphics[width=1in,height=1.25in,clip,keepaspectratio]{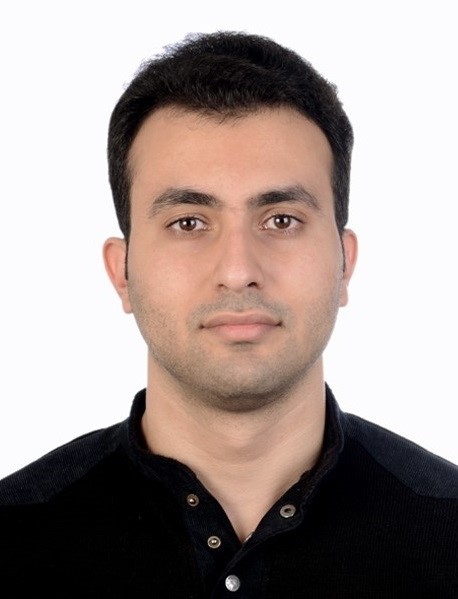}}]
{Ramin Zarei Sabzevar} received the B.Sc. and M.Sc. degrees in computer engineering and artificial intelligence from Ferdowsi University of Mashhad, Mashhad, Iran in 2014 and 2017 respectively. He is currently a research assistant and visiting lecturer at computer engineering department, Ferdowsi University of Mashhad.
He is also a lecturer at computer engineering department in Sadjad University of Technology, Mahhad, Iran.
Ramin's main research interests include neural networks, deep learning, information theory and probabilistic graphical models.
\end{IEEEbiography}
\begin{IEEEbiography}
[{\includegraphics[width=1in,height=1.25in,clip,keepaspectratio]{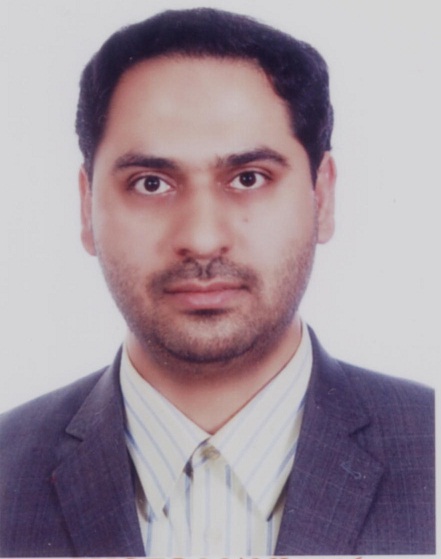}}]
{Kamaledin Ghiasi-Shirazi} received the diploma in mathematics and physics from Alborz high school, Tehran, Iran in 1997, the B.Sc. degree in computer software engineering from Shahid Beheshti University, Tehran, Iran in 2001, the M.Sc. degree in artificial intelligence from Sharif University of Technology, Tehran, Iran in 2004, and the Ph.D. degree in artificial intelligence from Amirkabir University of Technology, Tehran, Iran in 2010.  He has been an assistant professor of computer engineering department at Ferdowsi University of Mashhad, Mashhad, Iran, since 2012. His current research interests include neural networks, deep learning, pattern recognition, machine learning, kernel methods, and probabilistic graphical models.
\end{IEEEbiography}
\begin{IEEEbiography}
[{\includegraphics[width=1in,height=1.25in,clip,keepaspectratio]{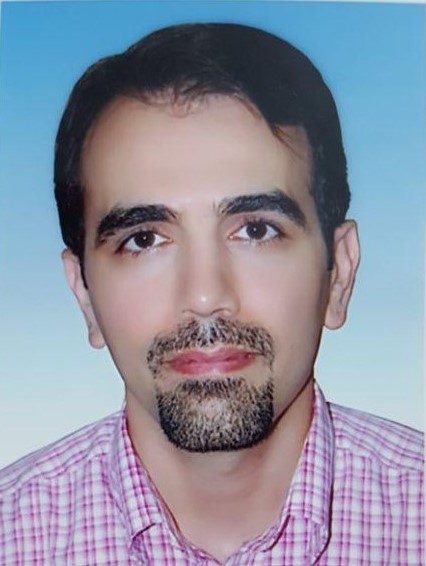}}]
{Ahad Harati} received his BSc. and MSc. degrees in computer engineering and AI \& Robotics from Amirkabir University of Technology and Tehran University, Tehran, Iran in 2000 and 2003 respectively.
He was awarded with Ph.D. in Manufacturing Systems \& Robotics from Swiss Federal Institute of Technology (ETHZ), Zurich, Switzerland, in 2008. He is currently an Assistant Professor in Computer Engineering department, Ferdowsi University of Mashhad. His areas of research is Machine Learning, Probabilistic Models, and Robot Perception.
\end{IEEEbiography}
\end{document}